\documentclass[runningheads]{llncs}

\usepackage{eccv}

\usepackage{eccvabbrv}

\usepackage{graphicx}
\usepackage{booktabs}

\usepackage[accsupp]{axessibility}  

\usepackage[pagebackref,breaklinks,colorlinks,citecolor=eccvblue]{hyperref}

\usepackage{orcidlink}

\usepackage{multirow}
\usepackage{amsmath}
\usepackage{graphicx}

\begin{document}

\title{BlazeBVD: Make Scale-Time Equalization Great Again for Blind Video Deflickering} 

\titlerunning{ }

\author{Xinmin Qiu \inst{1} \and
Congying Han\inst{1}  \and
Zicheng Zhang\inst{1} \and
Bonan Li\inst{1}  \and
Tiande Guo\inst{1}  \and
Pingyu Wang\inst{2} \and
Xuecheng Nie\inst{3}
}


\institute{University of Chinese Academy of Sciences, Beijing, China\\ \and
Sichuan University, Sichuan, China  \and MT Lab, Meitu Inc., Beijing, China
}

\maketitle
\begin{figure}
\vspace{-6mm}
\begin{center}
    \includegraphics[width=1.0\linewidth]{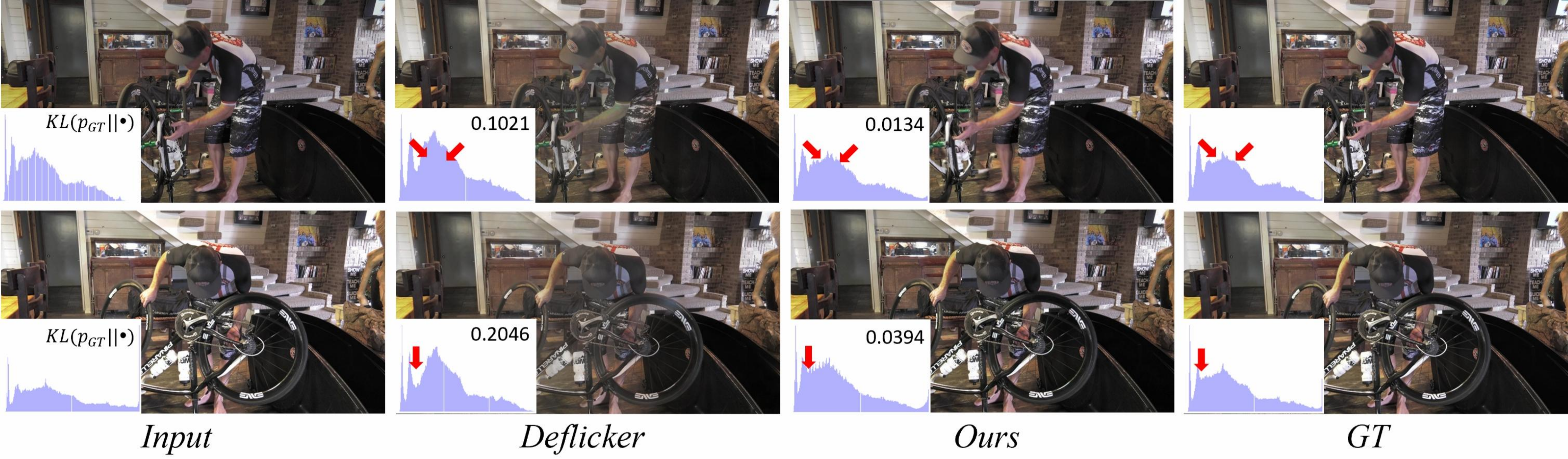}
\end{center}
   \vspace{-5mm}
   \caption{\textbf{Comparisons of the proposed \textit{BlazeBVD}}. We present flickering input, GT, Deflicker~\cite{Lei_2023} and our BlazeBVD processed video frames, and illumination histograms along with KL divergence about GroundTruth. Our method recovers the illumination histograms well while avoiding the appearance of color artifacts and color distortions (such as the man's arm in the second column). Better see in color with 2$\times$ zoom.}
   \vspace{-6mm}
\label{tear}
\vspace{-2mm}
\end{figure}
\vspace{-6mm}

\begin{abstract}
Developing blind video deflickering (BVD) algorithms to enhance video temporal consistency, is gaining importance amid the flourish of image processing and video generation. 
However, the intricate nature of video data complicates the training of deep learning methods, leading to high resource consumption and instability, notably under severe lighting flicker. 
This underscores the critical need for a compact representation beyond pixel values to advance BVD research and applications. 
Inspired by the classic scale-time equalization (STE), our work introduces the histogram-assisted solution, called \textbf{\textit{BlazeBVD}}, for high-fidelity and rapid BVD. 
Compared with STE, which directly corrects pixel values by temporally smoothing color histograms, BlazeBVD leverages smoothed illumination histograms within STE filtering to ease the challenge of learning temporal data using neural networks. 
In technique, BlazeBVD begins by condensing pixel values into illumination histograms that precisely capture flickering and local exposure variations.  These histograms are then smoothed to produce singular frames set, filtered illumination maps, and exposure maps. Resorting to these deflickering priors, BlazeBVD utilizes a 2D network to restore faithful and consistent texture impacted by lighting changes or localized exposure issues. 
BlazeBVD also incorporates a lightweight 3D network to amend slight temporal inconsistencies, avoiding the resource consumption issue. 
Comprehensive experiments on synthetic, real-world and generated videos, showcase the superior qualitative and quantitative results of BlazeBVD, achieving inference speeds up to 10$\times$ faster than state-of-the-arts. 
\end{abstract}

\section{Introduction}\label{sec:intro}
As social applications and the multimedia industry continue to expand, videos have become an essential medium for conveying information in daily life~\cite{Pfeuffer_2019_temporal}. Achieving a high-quality video demands both image clarity and temporal consistency, nevertheless, video flickering frequently compromises the temporal consistency, often resulting from shooting environment and camera hardware limitations~\cite{Delon_Desolneux_2010, Kanj_Talbot_Luparello_2017}. This issue is further exacerbated when image processing techniques are applied to video frames. Moreover, flickering artifacts and color distortion are also prevalent in recent video generation technologies, including those based on Generative Adversarial Networks~(GANs)~\cite{Chu_2020_gan_video, Saito_Saito_Koyama_Kobayashi_2020_gan, Skorokhodov_2022_gan} and Diffusion Models~(DMs)~\cite{wu_2023_dm_video, Mei_Patel_2023_dm, wang2024magicvideov2_dm}. Since even certain types of flicker can significantly detract from the viewing experience, it is crucial to develop video deflickering techniques to eliminate flicker and preserve the integrity of video content given unknown degradation in various video processing scenarios.

Several seminal works~\cite{Lei_2023, Delon_2006_STE, Lai_2018} have addressed the blind video deflickering~(BVD) task, in which methods refer to that (1) is agnostic to flickering patterns or levels (\eg, old movies, high-speed cameras, processing artifacts and color distortions), and (2) operates on a single flicker video without requiring additional guidance (\eg, flicker types, reference videos). In other words, BVD methods are blind to flicker types and guidance, making them widely applicable. Given these, the BVD task is challenging due to the lack of additional prior information guidance. Previous methods have explored traditional filtering~\cite{Piti2003RemovingFF_tradition,Delon_2006_STE}, forced temporal consistency~\cite{Lai_2018,Lei_2020}, and atlas adjustment~\cite{Lei_2023}, \etc. 
Despite significant advancements by deep-learning methods in BVD task, several critical issues hinder their broader application. First, these approaches demand substantial resources during both the training and inference stages, notably manifesting in slow processing speeds for individual videos. Second, their effectiveness is compromised by two main factors: (i) partial color artifacts and distortion are present in the resulting video, causing differences in color fidelity and detail compared to the original video; (ii) Overexposure or underexposure due to lighting leads to significant loss of texture details and introduces new local flicker. 
These issues are largely attributed to the spatio-temporal complexity within video data, where the sheer volume of pixels presents a formidable obstacle for neural networks, including the advanced large vision models, to learn and maintain global visual consistency. Consequently, given that flicker fundamentally involves local or instantaneous shifts in illumination, effectively addressing the BVD task necessitates a representation more compact and adept at capturing illumination fluctuations than pixel values, whereas this aspect has not yet been explored by previous deep BVD methods.

In this paper, we draw inspiration from the classic flicker removal method scale-time equalization (STE) to facilitate deep BVD with histogram representation. An image histogram is defined as a distribution of pixel values, and it has been widely applied in image manipulation to adjust image brightness or contrast. Given an arbitrary video, STE temporally smooths histograms with Gaussian filtering and corrects pixel values in each frame with histogram equalization,  improving the visual stability of the video. Although STE is only effective for some mild flickers, it validates that 1) histogram, which is much more compact than pixel values, can well delineate the illumination and flicker information, and 2) the video with a smoothed histogram series is visually appealing without obvious flicker, as shown in~\cref{tear}. Therefore, it is promising to advance the quality and speed of deep BVD with the hints from STE and histogram.

We introduce Blaze Blind Video Deflickering, dubbed as BlazeBVD, which is a histogram-assisted approach to achieve fast and faithful texture restoration given illumination fluctuation and over-/under-exposure. Compared to previous deep methods, BlazeBVD is the first to leverage histogram meticulously to ease the learning complexity of BVD task. Specifically, BlazeBVD comprises three stages: At first, we introduce STE to rectify the histogram series of video frames under the illumination space, and extract \textit{deflickering priors} including singular frames set, filtered illumination maps, and exposure maps. Second, due to the stable temporal performance of filtered illumination maps, they will be used as prompt conditions of a global flicker removal module (GFRM) containing 2D spatial networks to guide the color correction of video frames. On the other hand, a local flicker removal module (LFRM) is based on the optical flow warp to restore local over-/under-exposure regions labeled by exposure maps. Finally, we introduce a lightweight spatio-temporal network to process all frames, where an adaptive mask weighted warping loss is designed to improve video coherence. Our contributions can be summarized as follows:

(1) We present BlazeBVD, a histogram-assisted blind video deflickering method that simplifies the complexity and resource consumption of learning video data. At its core, BlazeBVD utilizes deflickering priors from STE, including filtered illumination maps for guiding the elimination of global flicker, singular frames set for identify the indexes of flicker frames, and exposure maps to identify regions affected locally by over-/under-exposure.

(2) Leveraging deflickering priors, BlazeBVD incorporates a Global Flicker Removal Module (GFRM) and a Local Flicker Removal Module (LFRM). These modules work together to efficiently correct global illumination and locally exposed textures in individual adjacent frames, significantly reducing processing time compared to handling the entire video. For enhanced coherence across frames, a lightweight spatio-temporal network is also integrated, boosting performance without significant time consumption. 

(3) Through comprehensive experiments on synthetic, real-world and generated videos, we showcase the superior qualitative and quantitative results of BlazeBVD, achieving model inference speeds up to 10$\times$ faster than state-of-the-arts. To the best of our knowledge, BlazeBVD also represents the first method to effectively tackle both illumination fluctuations and exposure challenges.

\section{Related Work}
\paragraph{\textbf{Video deflickering}} addresses the degradation in temporal consistency~\cite{Xu_2020_videowork, perez2018perceptual_consistency, moniz2019luciddream_consistency, Thasarathan_2019_consistency, Park_2019_consistency} resulting from abnormal camera, lighting, exposure or image processing algorithms. The relevant methods can be divided into three main categories.  The first type of approach focuses on addressing specific flickering types, \textit{e.g.}, high-speed cameras~\cite{Kanj_Talbot_Luparello_2017}, old film~\cite{Delon_Desolneux_2010}. Delon~\cite{Delon_2006_STE} performs an axiomatic analysis of flicker, and proposed a fast method for global removal based on scale space theory to filter in color space directly.
 The second type of approach, blind temporal consistency~\cite{Zhou_2020_consistency, Thimonier_2021_consistency}, aims to generate temporally stable output videos for arbitrary videos regardless of flickering or other artifacts.
 Bonneel \etal~\cite{Bonneel_2015} calculate the gradient of the input frame as a guide to reduce the degree of inconsistency between frames. Lai \etal~\cite{Lai_2018} input two consecutive frames as guidance. Lei \etal~\cite{Lei_2020} directly learn the mapping function between input frames and processing frames to reach the purpose of inter-frame information consistency. Despite the generalization performance, such methods tend to be less efficient and hard to handle the complex flicker in practice. 
 The third type of approach is blind video deflickering.  Lei \etal~\cite{Lei_2023} propose to solve universal blind video flicker using the neural atlas and video consistency. Meanwhile, some commercial software can remove flicker by integrating various known-flicker methods, such as DEFlicker~\cite{revision} and Flicker Free~\cite{flicker_free}. However, most of these methods require users to have flickering type knowledge background or slow repair speed. Our approach does not require prior cognition of the specific information flashing, and achieves fast deflickering speeds so that most users can efficiently process more videos.
\paragraph{\textbf{Exposure correction.}} The purpose of exposure correction is to enhance an improperly exposed image to achieve a satisfactory visual effect. Information is lost due to over-/under-exposure input images. How to solve this problem intelligently and automatically, there have been many techniques in the image domain~\cite{ECLNet_2022_exposure, Ma_2022_CVPR_exposure, Huang_2022_ENC_exposure, Huang_2022_FECNet_exposure, Wang_2022_LCDPNet_exposure}. ERL~\cite{huang2023learning_exposure} connects the under-/over-exposed optimization processes by correlating and constraining the relationships in the mini-batch correction process. DA~\cite{Wang_2023_DA_exposure} proposes to decouple contrast enhancement and detail recovery in each convolution process and design detail perception units to be inserted into the existing CNN-based exposure correction network. However, there is no general approach regarding local exposure in video tasks.
\paragraph{\textbf{Optical flow}} finds the corresponding relationship between the previous frame and the current frame by using the changes of pixels in the time domain and the correlation between adjacent frames, to calculate the motion information of objects between consecutive frames~\cite{dosovitskiy2015flownet, ilg2017flownet2, teed_2020_raft, eslami_2024_rethinking_raft}. In recent years, it has been used in video-related visual tasks~\cite{zhan2019_opticalflow}. The main target of flow-based methods is to produce accurate estimates of the optical flow through different network architectures, such as utilizing bidirectional optical flow estimation and refining inter-frame coherence.






\vspace{-1mm}
\section{Preliminaries}
In this section, we introduce basic definitions and notations, and review relevant techniques involving image histograms for use in BlazeBVD.

\paragraph{\textbf{Image histogram}} is a graphical representation illustrating the distribution of pixels across different intensity levels within an image~\cite{Bassiou_Kotropoulos_2007_histogram}. For a single-channel image $I$ with height $H$ and width $W$, its histogram $\mathcal{H}$ can be calculated as
\begin{equation}\label{Histogram_formula}
        \mathcal{H}(\lambda)=\frac{n_\lambda}{HW}, \hspace{4pt} if \ \lambda \in I \ otherwise\  0,
\end{equation}
where $n_\lambda$ represents the count of pixels having the intensity $\lambda$.

\paragraph{\textbf{Histogram matching}} or equalization~\cite{Bottenus_Byram_Hyun_2021_matching, Xu_2023_histogram, afifi2021_histogram, zhang2021_histogram}, is a technique to adjust the contrast of an image by transforming its histogram towards a predefined distribution.  The cumulative histogram of an image histogram $hist$ is defined by
    $Hist(\lambda)=\Sigma_{r=0}^\lambda \mathcal{H}(r).$
Give a target histogram $\mathcal{H}'$ and its cumulative histogram $Hist'(\lambda)$, the histogram matching process can be described as  
\begin{equation}
    \begin{aligned}
       \mathcal{Q}(\lambda;\mathcal{H}, \mathcal{H}')=Hist'^{-1}(Hist(\lambda)),\hspace{2pt} \forall\lambda \in  I,
    \end{aligned}
    \label{Histogram_matching}
\end{equation}
This procedure yields an image $\tilde{X}$, whose histogram closely aligns with $Hist'$.

\paragraph{\textbf{Scale-Time Equalization}} (STE) is a filtering approach~\cite{Delon_2006_STE} to adjust the intensity distribution across both spatial and temporal dimensions, mitigating flickers encountered in video sequences. The classic scale-space theory posits that convolving an image with a Gaussian kernel results in smoothing and the loss of image structure. Intuitively, while STE is based on scale-space theory, it applies Gaussian smoothing to sequences of histograms instead of the images.  Given a flickering video sequence $\{X_t\}_{t=1}^{T}$, leveraging \cref{Histogram_formula} can get the histogram sequence $\{\mathcal{H}_t\}_{t=1}^{T}$. For arbitrary time index $t$ and intensity value $\lambda$, STE can be formulated as the convolution:
\begin{equation}
    \begin{aligned}
        STE(\lambda,t) &= G_{s}(\tau) * \mathcal{Q}(\lambda;\mathcal{H}_t, \mathcal{H}_{\tau})
        \approx\sum_{\tau=t-l}^{t+l}G_{s}(\tau-t) \mathcal{Q}(\lambda;\mathcal{H}_t, \mathcal{H}_{\tau}),
    \end{aligned}
    \label{STE}
\end{equation}
where $l$ denotes the window radius, $G_s(t)=\frac{1}{\sqrt{4\pi s}}e^{-\frac{t^2}{4s}}$.
After STE, we will get a sequence of histograms  $\{\tilde{\mathcal{H}}_t\}_{t=1}^{T}$ and images $\{\tilde{X}_t\}_{t=1}^{T}$. Since histogram transformations do not create or cancel any image content, STE can achieve deflickering results without compromising the integrity of the original video. However, this also indicates that STE cannot compensate for the lost texture in flickering regions and produces vivid effects, which impedes its practical application.


\section{Methodology}
 Let $\{X_t\}_{t=1}^{T}$ represent a video sequence affected by flickering defects and $\{G_t\}_{t=1}^{T}$ represent the groundtruth clean video. In this section, we propose \textit{BlazeBVD} as a universal solution to eliminate these flicker and produce a temporally consistent video $\{O_t\}_{t=1}^{T}$. It is important to note that the types of flicker addressed in our work can vary widely, involving long-/short-term flicker in color and brightness.

 Distinguished from existing prior BVD approaches, BlazeBVD employs a histogram-assisted approach, leveraging deflickering priors from Scale-Time Equalization (STE)~\cite{Delon_2006_STE} to streamline the complexity and resource demands of BVD task. Beyond ensuring stable temporal performance, BlazeBVD stands out as the first method capable of faithfully restoring the color and texture affected by illumination fluctuations and exposure challenges, including both over-/under-exposure.

\begin{figure}[tb]
\begin{center}
    \includegraphics[width=1.0\linewidth]{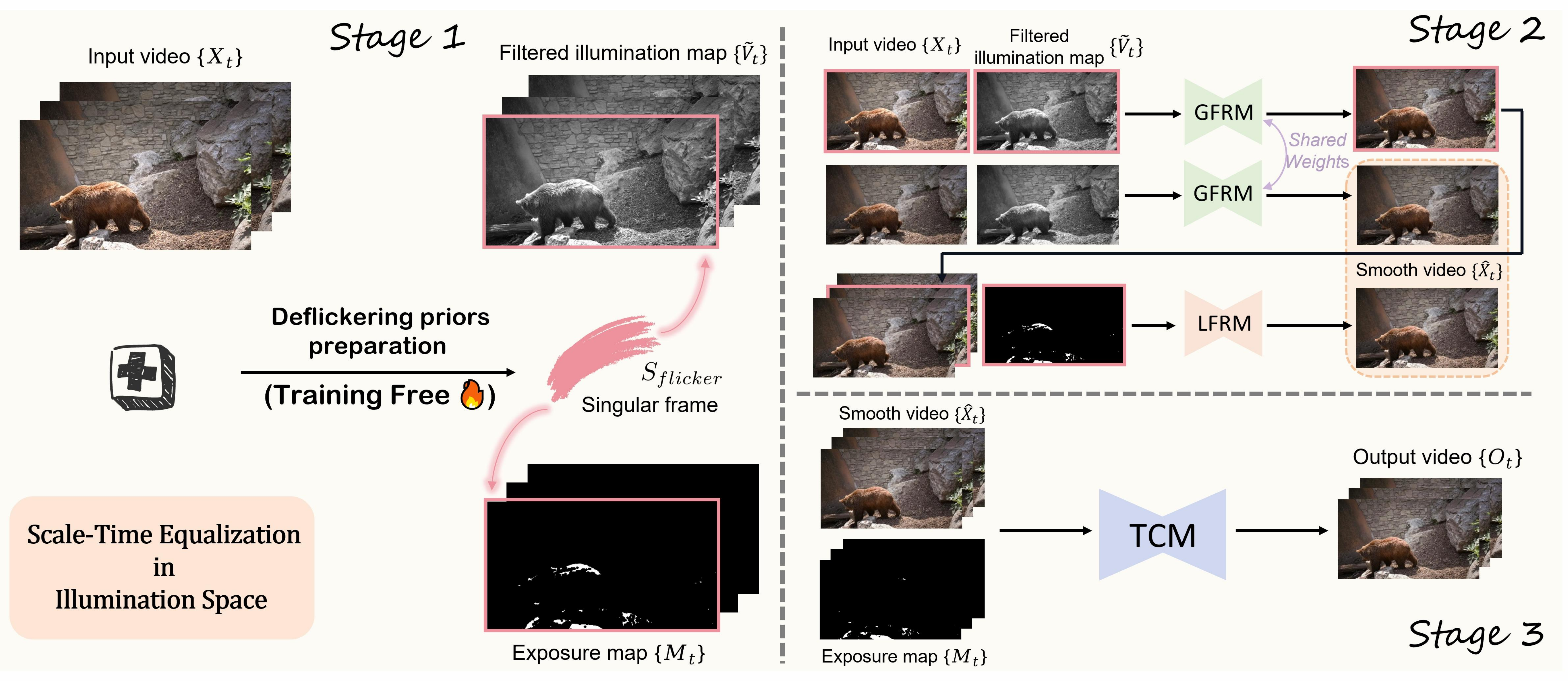}
\end{center}
    \vspace{-5mm}
   \caption{\textbf{The framework of our approach \textit{BlazeBVD}}. We first extract flicker prior information about the input video and correct the brightness representation with STE-assisted histogram filtering in illuminance space (Stage1). Then, the prior is leveraged to remove temporal flicker and over-/under-exposure flicker from the global and local perspectives (GFRM and LFRM in Stage2). Finally, the temporal consistency of the processed video is improved (TCM in Stage3).}
\label{frame_work}
\vspace{-5mm}
\end{figure}

\subsection{Overview of BlazeBVD}
As illustrated in~\cref{frame_work}, the BlazeBVD pipeline is structured into three main stages to address video deflickering challenges effectively: 
(1) Deflickering priors preparation. This initial stage leverages STE to correct the histogram within an illumination space representation. Key outputs from this stage include the filtered illumination map $\{\tilde{V}_t\}_{t=1}^{T}$,  identifying the index set of singular frames $S_{flicker}$, and generating exposure maps $\{M_t\}_{t=1}^{T}$. These elements serve as crucial prior information, guiding the deflickering process in subsequent stages.
(2) Global and local flicker removal. The second stage involves two flicker removal strategies. A global flicker removal module utilizes the filtered illumination map $\{\tilde{V}_t\}_{t=1}^{T}$ to address temporal flicker issues across the video. A local over-/under-exposed region flicker removal module instructed by the singular frames set $S_{flicker}$ and exposure maps $\{M_t\}_{t=1}^{T}$, restores the texture in the over-/under-exposure region to compensate for the lost high-frequency details.
(3) Adaptive temporal consistency. The final stage introduces a spatio-temporal network under the guidance of an adaptive warping loss to improve video coherence and temporal consistency. 

These stages delineated in the following~\cref{sec::stage1},~\cref{sec::stage2}, and~\cref{sec::stage3}, together provide a comprehensive solution to enhance the visual quality and temporal consistency of video content. 

\subsection{Stage1: Deflickering priors preparation} \label{sec::stage1}
\begin{figure}[tb]
\begin{center}
    \includegraphics[width=1.0\linewidth]{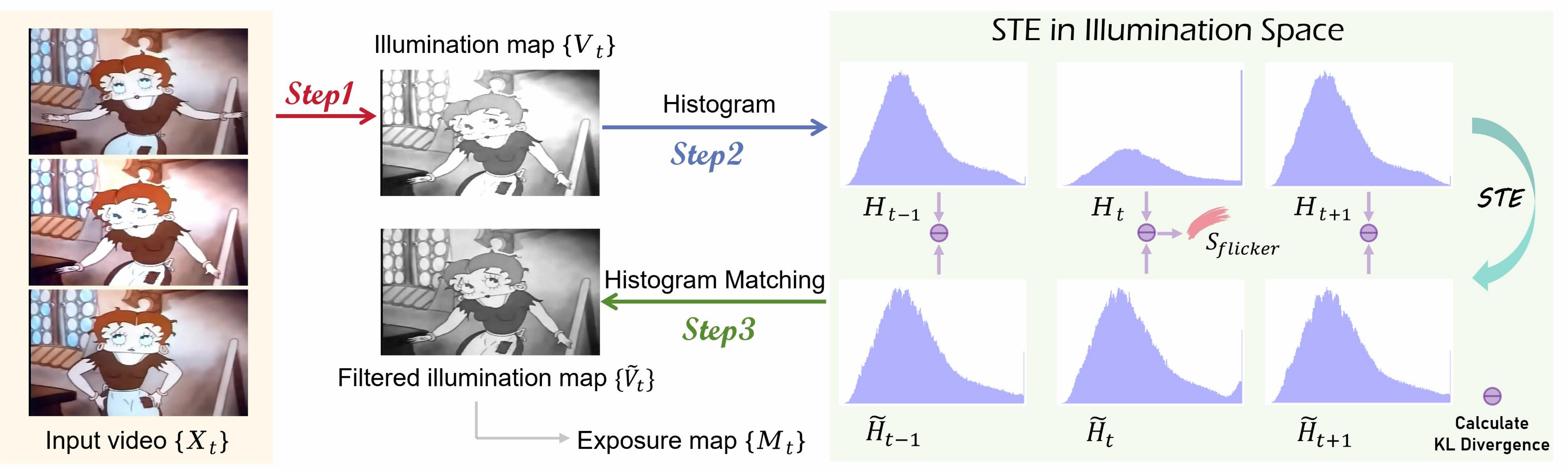}
\end{center}
    \vspace{-5mm}
   \caption{\textbf{Pipeline of Stage1}.  The illumination maps are corrected in the temporal dimension by STE, which alleviates the difficulty of subsequent temporal data learning using networks. The deflickering priors can be extracted: $\tilde{V}_t$, $M_t$ and $S_{flicker}$.}
\label{stage1_frame}
\vspace{-5mm}
\end{figure}

\paragraph{\textbf{Spatio-temporal challenge}.} An effective BVD model should strategically avoid the intricacies of spatio-temporal complexity across pixels, nevertheless, current deep learning approaches to BVD have not adequately addressed this challenge.
For instance, the state-of-the-art BVD method,   Deflicker~\cite{Lei_2023}, employs a neural atlas~\cite{kasten2021_atlas} as a unified representation of the entire video. While this can offer consistent guidance for the deflickering process, the effectiveness of neural atlases significantly depends on the video duration and the network capacity. Processing an 8-second video to create an atlas can take around 10 minutes, and longer videos exacerbate the demands on time, memory, and computational resources.  Consequently, there is a further need for research into more compact representations that can facilitate deflickering without these limitations.


\paragraph{\textbf{Illumination histogram.}} We introduce histogram representation~\cite{Bassiou_Kotropoulos_2007_histogram}, which utilizes statistical distribution to illustrate the global properties of image pixel values. As illustrated in \cref{stage1_frame}, we first extract the illumination maps $\{V_t\}_{t=1}^{T}$ according to the conversion between HSV and RGB color space:
\begin{equation}
    V_t=\max_{R,G,B\in X_t}\{R,G,B\},\ t\in \{1,2,...,T\}.
\end{equation}
Next, by applying~\cref{Histogram_formula} to each illumination map, we utilize illumination histograms $\{H_t\}_{t=1}^{T}$ to represent the illumination characteristics of video frames. These illumination histograms reflect important flickering information: As shown in~\cref{stage1_frame}, frames without distinct illumination changes have similar histograms, while the histogram of a flicker frame has a distribution distinguished from others. Furthermore, spikes in the head or tail area of the distribution indicate potential over-/under-exposure problems in the image. To this end, reasonable utilization of histogram information is significant for flicker elimination and alleviation of spatio-temporal complexity.


\paragraph{\textbf{Deflickering priors}.} We introduce deflickering priors extracted from illumination histograms to facilitate the deflickering process. Initially, considering that the illumination histograms of natural videos exhibit continuous changes in the temporal dimension, we utilize Scale-Time Equalization (STE) as defined in \cref{STE} to smooth the illumination histograms, as illustrated in \cref{stage1_frame}. Through histogram matching, we further obtain filtered illumination maps $\{\tilde{V}\}_{t=1}^{T}$ that exhibit no obvious flicker compared to the original ones, thereby enabling the subsequent independent processing of each frame by the neural network.
Secondly, we identify the singular frames set $S_{flicker}$ containing the indexes of flicker frames, based on the divergence of histograms before and after STE:
\begin{equation}
    \begin{aligned}
        S_{flicker}=\{t\in[1,T]|
        \text{KL}(\tilde{H}_t\parallel H_t) > \bar{\epsilon}_{t} \},
        \ \bar{\epsilon}_{t} = \frac{1}{2n+1}\sum_{\tau=t-n}^{t+n}\text{KL}(\tilde{H}_\tau \parallel H_\tau),
    \end{aligned}
    \label{S_flicker}
\end{equation}
where $n$ is the moving average radius.
Finally, as shown in \cref{frame_work}, we extract the exposure maps $\{M_t \in\mathbb{R}^{H \times W}\}_{t=1}^{T}$ of local over-/under-exposure according to the filtered illumination map: 
\begin{equation}
    \begin{aligned}
        M_t(i,j) =
        \begin{cases}
            1 & \tilde{V}(i,j)<\epsilon_1\ \text{or}\ \tilde{V}(i,j)>\epsilon_2,\\
            0 & \text{otherwise}.
        \end{cases}
    \end{aligned}
    \label{mask}
\end{equation}
Here, $\epsilon_1$ and $\epsilon_2$ are darkness and exposure thresholds, respectively, and we consider illumination values belonging to $[\epsilon_1, \epsilon_2]$ as not posing an exposure risk.  

In summary, we leverage illumination histograms to prepare deflickering priors including filtered illumination maps $\{\tilde{V}\}_{t=1}^{T}$, exposure maps $\{M_t\}_{t=1}^{T}$, and the singular frames set $S_{flicker}$, for alleviating the spatio-temporal complexity and enhancing deflickering in the subsequent learning stages.


\subsection{Stage2: Global and local flicker removal} \label{sec::stage2}
In this stage, simple 2D spatial neural networks are integrated with deflickering priors to eliminate both global and local flickering flaws. Here, “global” refers to obvious temporal inconsistency due to illumination fluctuation, while “local” refers to lost texture details due to exposure and darkness. This approach allows for the direct generation of high-quality videos from flicker frames, regardless of the spatio-temporal complexity.

\paragraph{\textbf{Global deflickering}.} BlazeBVD is equipped with a global flicker removal module (GFRM), which incorporates a 2D-Unet to correct each frame $X_t$ with the guidance of the filtered illumination map $\tilde{V}_t$. In other words, the corrected image $\tilde{X}_t=Unet_{\theta}(X_t, \tilde{V}_t)$, where $\theta$ denotes the parameters of the 2D-Unet. The network is trained on the BVD dataset $\mathcal{D}$ by minimizing the L2 loss function between the predicted images and the groundtruth images:
\begin{equation}
    \begin{aligned}
        \mathcal{L}_t =\mathbb{E}_{(X_t, G_t)\in \mathcal{D}} \| Unet_\theta(X_t, \tilde{V}_t) - G_t \| .
    \end{aligned}
    \label{spatial unet loss}
\end{equation}
\paragraph{\textbf{Local deflickering}.} BlazeBVD is also equipped with a local flicker removal module~(LFRM) to enhance local texture details. As depicted in \cref{Ablation_figure}, over-exposure can transform intricate texture in local regions into uniform pure, while local under-exposure can result in black areas. This loss of detailed texture cannot be compensated for by previous BVD methods, such as Deflicker~\cite{Lei_2023}. We leverage LFRM to transfer detailed information from neighboring frames to these ill regions. Given the corrected singular frame $\tilde{X}_t$ in $S_{flicker}$ with local exposure, along with the previous frame $\tilde{X}_{t-1}$ and the next frame $\tilde{X}_{t+1}$, we first compute the optical flow to extract part of the detailed texture:
\begin{equation}
    \begin{aligned}
        o_{\tau \rightarrow t}=Flow(X_{\tau}, X_t),\ \tau \in \{t+1,t-1\} \ \text{and}\   t\in S_{flicker}.
    \end{aligned}
    \label{optical flow}
\end{equation}
 Here $Flow$ is the optical flow estimation function, in our work we set it as the popular RAFT model~\cite{teed_2020_raft,eslami_2024_rethinking_raft}. We migrate the possible detail textures of the preceding and following frames into the current frame:
\begin{equation}
    \begin{aligned}
        \hat{X}_t = \mathcal{F}(&M_t \odot \mathcal{W}(\tilde{X}_{t-1},  o_{t-1\rightarrow t}) + (1-M_t) \odot \tilde{X}_t,\tilde{X}_t, \\&M_t \odot \mathcal{W}(\tilde{X}_{t-1},  o_{t+1\rightarrow t}) + (1-M_t) \odot \tilde{X}_t),t\in S_{flicker},
    \end{aligned}
    \label{exposure add}
\end{equation}
where $\mathcal{W}$ is the warping operator and $\mathcal{F}$ is the fusion network from~\cite{Wang_2022_LCDPNet_exposure}.
\vspace{-3mm}

\subsection{Stage3: Adaptive temporal consistency} \label{sec::stage3}
The former stages have well addressed the main illumination fluctuation and exposure problems in BVD task. Nevertheless, a video may also have content inconsistencies to largely degrade the visual effect, and the former stage may also cause this.  Therefore, We introduce the Temporal Consistency Model (TCM) to refine video $\{\hat{X}_t\}_{t=1}^{T}$ as the final outputs $\{O_t\}_{t=1}^{T}$. TCM is developed based on the architecture by RTN~\cite{Wan_2022_CVPR_videorestore}, but with a temporal consistency loss of adaptive masks to refine local artifacts:
\begin{equation}
    \begin{aligned}
E'_{\text {pair }}\left(O_t, O_s\right) & = \left\| W_t \odot (M_t+1) \odot M_{t, s}\odot \left(O_t -\mathcal{W}\left(O_s\right)\right) \right\|_1, \\
\mathcal{L}_{warp}& =\frac{1}{T-1} \sum_{t=2}^T\left\{E'_{\text {pair}}\left(O_t, O_1\right)+E'_{p a i r}\left(O_t, O_{t-1}\right)\right\},
    \end{aligned}
    \label{warp loss}
\end{equation}
where $W_t$ is the loss weight for the over-/under-exposure area and $M_t$ is the mask obtained from the former stage. $\mathcal{W}$ is the warping operator and the frame after warping is computed by optical flow estimation~\cite{teed_2020_raft, eslami_2024_rethinking_raft}. $M_{t, s}$ represents the mask after optical flow estimation.
This loss avoids the video blurring problem introduced in Deflicker~\cite{Lei_2023}. The training loss of TCM consists of reconstruction loss $\mathcal{L}_{rec}$, perceptual loss $\mathcal{L}_{per}$~\cite{Johnson_2016_VGG}, spatio-temporal adversarial loss $\mathcal{L}_{adv}$~\cite{Chang_2019_ICCV_adv_loss} and designed adaptive weighted warping loss $\mathcal{L}_{warp}$. Combining all the above losses, the overall objective to optimize TCM is:
\begin{equation}
    \begin{aligned}
        \mathcal{L}_{TCM} = \alpha_1 * \mathcal{L}_{rec} + \alpha_2 * \mathcal{L}_{per} + \alpha_3 * \mathcal{L}_{adv} + \alpha_4 * \mathcal{L}_{warp}. \\
    \end{aligned}
    \label{loss stage3}
\end{equation}
In our work, we set $\alpha_1=1.0$, $\alpha_2=1.0$, $\alpha_3=0.01$ and $\alpha_4=0.1$.

\vspace{-5mm}
\section{Experiments}
\vspace{-1mm}
\subsection{Datasets}\label{sec::datasets}

We inherit the training dataset DAVIS-2017-Train, MS-COCO and the testing dataset \textit{Blind Deflickering Dataset} from Deflicker~\cite{Lei_2023}, including Synthetic videos from DAVIS-2017-Val, Real-world videos and Generation videos. 

\noindent\textit{\textbf{Training dataset}.}
MS-COCO~\cite{Lin_2014_COCO} contains 118287 images in the train set and is used for the pre-training of Unet in GFRM.
DAVIS-2017-Train~\cite{Perazzi_2016, Pont-Tuset_arXiv_2017_DAVIS, Lai_2018} consists of 60 sequences with a total of 4219 labeled frames, with a video frame rate of 24fps and a resolution of 480p.
We create the synthetic process that provides the underlying facts for quantitative analysis. The training dataset is degraded by performing flicker on DAVIS-2017-Train, as shown in~\cref{formula_degradation}: 
\begin{equation}
    \begin{aligned}
        X_t=G_t+F_t,\ t\in \{1,2,...,T\},
    \end{aligned}
    \label{formula_degradation}
\end{equation}
where $\{G_t\}_{t=1}^{T}$ are clean video frames and $\{F_t\}_{t=1}^{T}$ are random synthesized flickering artifacts. The window size $W$ in $F_t$ has been set to denote the number of frames that share the same flickering artifacts, random sampling between [2, 12]. The value of $W$ controls the degree of flicker in the short- and long-term of the video.

\noindent\textit{\textbf{Testing dataset}.}
We conduct Synthetic videos, Real-world videos and Generation videos respectively, from the \textit{Blind Deflickering Dataset}~\cite{Lei_2023} and DAVIS-2017-Test processed by~\cref{formula_degradation}. Among them, Synthetic videos and DAVIS-2017-Test have ground truth and the others do not. We can only use no-referenced metric and qualitative observation. Generation videos refer to the processed videos generated by generative algorithms, which may lead to flicker artifacts due to the imperfections of the algorithms. Real-world videos including old movies, old anime, slow-motion, \etc are directly captured, unprocessed videos in the real world. 
\begin{figure}[tb]
\begin{center}
    \includegraphics[width=1.0\linewidth]{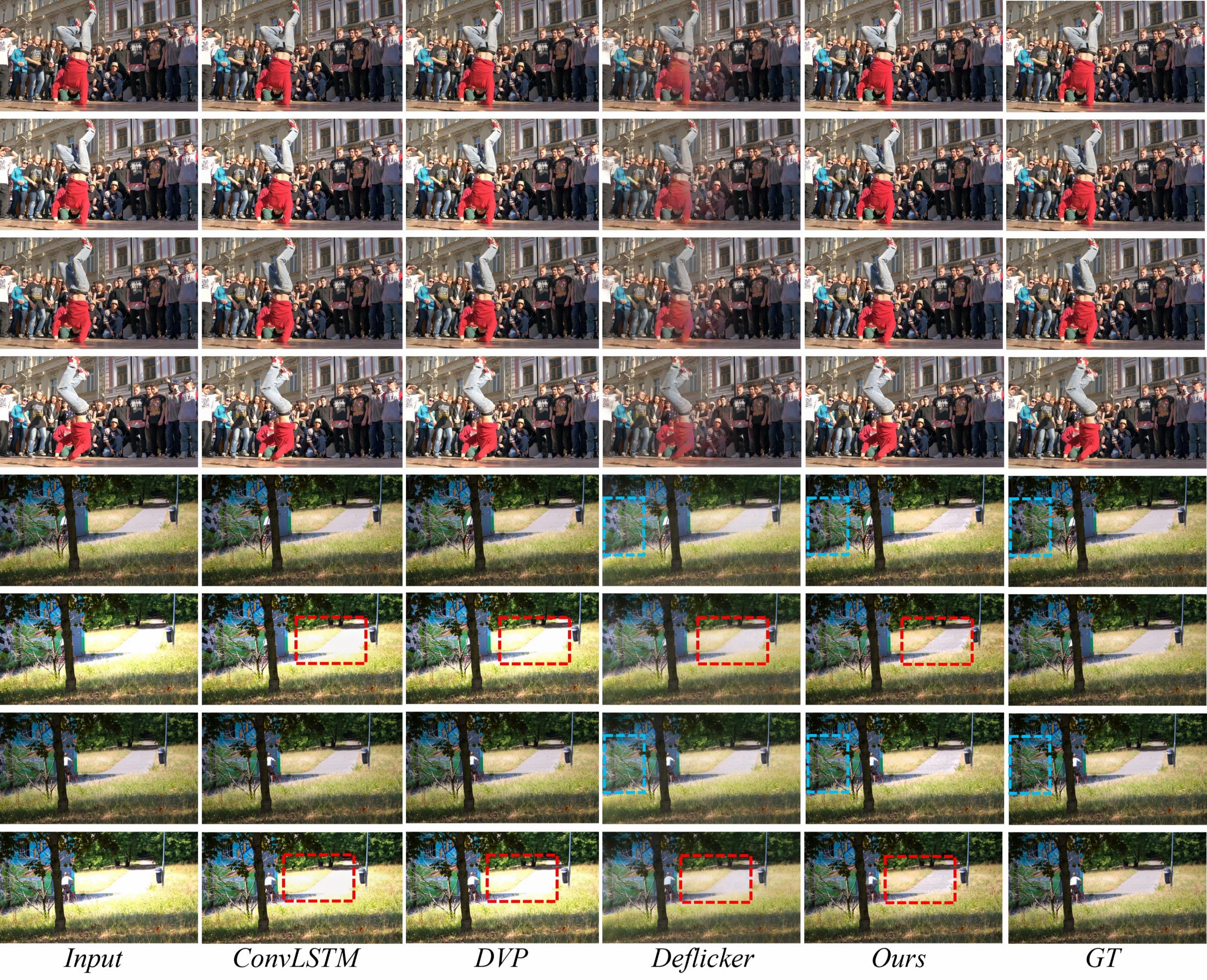}
\end{center}
   \vspace{-6mm}
   \caption{\textbf{Qualitative comparisons between previous methods and our \textit{BlazeBVD}.} Our method removes flicker and restores details in over-exposed regions (\textit{\textcolor{red}{red boxes}} in row 6 and row 8) while avoiding color artifacts (row 2 and row 4). Besides, \textit{BlazeBVD} also ensures the fidelity of the video content and avoids color distortion (\textit{\textcolor{cyan}{cyan boxes}} in row 5 and row 7). Zoom in for the best view and we recommend watching videos in the supplementary materials.}
   \vspace{-5mm}
\label{Qualitative1}
\end{figure}

\noindent\textit{\textbf{Comparison methods}.}
We compare our BlazeBVD with other 4 methods, including STE~\cite{Delon_2006_STE}, ConvLSTM~\cite{Lai_2018}, DVP~\cite{Lei_2020} and Deflicker~\cite{Lei_2023}. We set Deflicker as the baseline, since it is the first publicly presented method for BVD task and is currently the most advanced approach for general videos.
\vspace{-3mm}
\subsection{Comparisons with State-of-the-art Methods}\label{sec::comparison}
\noindent\textit{\textbf{Metrics}.}
We quantitatively compare the differences between our method and others with reference and no-reference metrics, including PSNR, SSIM~\cite{SSIM}, and $E_{warp}$~\cite{Lei_2020}. Among them, $E_{warp}$ is a no-reference metric used in videos. For each frame $O_t$, we calculate the warping error with frame $O_{t-1}$ and the first frame $O_1$ for considering both short-term and long-term consistency. The final $E_{warp}$ is calculated by:
\begin{equation}
    \begin{aligned}
E_{\text {pair }}\left(O_t, O_s\right) & = \left\| M_{t, s}\odot \left(O_t -\mathcal{W}\left(O_s\right)\right) \right\|_1, \\
E_{\text {warp }}\left(\left\{O_t\right\}_{t=1}^T\right) & =\frac{1}{T-1} \sum_{t=2}^T\left\{E_{\text {pair }}\left(O_t, O_1\right)+E_{p a i r}\left(O_t, O_{t-1}\right)\right\},
    \end{aligned}
    \label{warping metric}
\end{equation}
where $\mathcal{W}(\cdot)$ is the warping model and the frame after warping is computed by optical flow estimation. $M_{t, s}$ represents the mask after optical flow estimation, and the effective coordinate region after warping is recorded.

\paragraph{\textbf{Quantitative results}.}
We compare our BlazeBVD with the state-of-the-art method Deflicker~\cite{Lei_2023}, and other consistency methods.~\Cref{tab:DAVIS_results} and~\Cref{tab:synthetic_results} provide the comparative experimental results of various methods on DAVIS-2017-Test and Synthetic videos, where PSNR and SSIM of the resulting video are higher than those of the other methods. This shows that BlazeBVD contributes to the fidelity maintenance of video content, as well as the correctness of the missing details complement. The warping error of our method is a little higher than Deflicker, but it is closer to the ground truth. Through analysis, we find that warping error has inaccuracy in the estimation of optical flow, and this error is masked by image blurring in Deflicker. Therefore, we believe that $E_{warp}$, which has the same magnitude as baseline and GT, is the valid experimental comparison result. In~\cref{tab:realdata_results}, we compare BlazeBVD and Deflicker on Real-world videos and Generation videos. Since the videos in these two datasets with no reference are inherently discontinuous, where $E_{warp}$ cannot directly represent video continuity, we present the comparative results of user studies. To evaluate the perceived preference between the baseline Deflicker and the proposed method, each user needs to select a video with better perceptual quality between these two choices. In total, there are 50 users and 50 pairs in 7 datasets of comparisons. 

\begin{table}[tb]
{
\caption{\textbf{Quantitative comparisons} of synthetic videos in DAVIS-2017-Test.  Our method performs best in terms of PSNR and SSIM and $E_{warp}$ of all the methods.}
\vspace{-2mm}
  \label{tab:DAVIS_results}
  \setlength\tabcolsep{6pt}
  \centering
  \begin{tabular}{@{}lccccccc@{}}
    \toprule
    Method & RawVideo & ConvLSTM & DVP & STE & Deflicker & Ours & GT\\
    \midrule
    PSNR$\uparrow$       & -      & 24.110 & 24.136 & 26.138 & 23.932 & \textbf{28.609} & -\\
    SSIM$\uparrow$       & -      & 0.9256 & 0.9263 & 0.9321 & 0.9243 & \textbf{0.9638} & -\\
    $E_{warp}\downarrow$ & 0.1429 & 0.1369 & 0.1415 & 0.1117 & 0.0840 & \textbf{0.0825} & 0.1097\\
  \bottomrule
  \end{tabular}
}
\vspace{-2mm}
\end{table}

\begin{table}[tb]
{
\caption{\textbf{Quantitative comparisons in Synthetic videos from \textit{Blind Deflickering Dataset}}. $W$ is the window size and represents the number of frames that share the same flicker artifact. $L$ denotes the local seed window size.}
\vspace{-3mm}
  \label{tab:synthetic_results}
  \setlength\tabcolsep{3.5pt}
  \centering
  \begin{tabular}{@{}lcccccccc@{}}
    \toprule
    
    \multirow{2}{*}{VideoType} & \multicolumn{4}{c}{$E_{warp}\downarrow$} & \multicolumn{2}{c}{PSNR$\uparrow$}  & \multicolumn{2}{c}{SSIM$\uparrow$} \\
      & Raw & Deflicker & Ours & GT & Deflicker & Ours & Deflicker & Ours \\
    \midrule
    Synthetic  &&&&&&&&  \\
    \quad -$W=1$   & 0.1933 & 0.0848 & \underline{0.0895} & 0.1056 & 25.9312 & \textbf{29.6738} &0.9403 & \textbf{0.9539}\\
    \quad -$W=3$   & 0.1522 & 0.0844 & \underline{0.0890} & 0.1056 & 25.4996 & \textbf{28.8272} &0.9380 & \textbf{0.9555}\\
    \quad -$W=10$  & 0.1307 & 0.0855 & \underline{0.0904} & 0.1056 & 24.3842 & \textbf{26.9966} &0.9344 & \textbf{0.9543}\\
    \quad -\ $L=3$  & 0.1113 & 0.0875 & \underline{0.0900} & 0.1056 & 25.3874 & \textbf{27.0881} &0.9271 & \textbf{0.9517}\\
    \midrule
    Average & 0.1469 & 0.0856 & \underline{0.0897} & 0.1056 & 25.3006 & \textbf{28.1464} &0.9349 & \textbf{0.9538}\\
    
  \bottomrule
  \end{tabular}
  \vspace{-4mm}
}
\end{table}



\paragraph{\textbf{Qualitative results}.}
Both ConvLSTM and DVP are temporal consistency models that require a reference, importing the input video as the reference, and the processed video cannot remove significant flicker, as shown in~\cref{Qualitative1}. Additionally, the flicker caused by over-/under-exposure is prone to lose the local texture (column 1 in~\cref{Qualitative1}). The atlas-based representation in Deflicker cannot accurately locate these points with different local features, resulting in the loss of all details on the processed video. Instead, our method designs local positioning and adjacent frame details transfer in LFRM to preserve part of the texture on the road, as shown in the red box of~\cref{Qualitative1}. We also reveal the color artifact in column 4 of~\cref{Qualitative1}          that appears at baseline, due to the excessive weight of warping error in the training loss when refining temporal consistency. Therefore, our designed adaptive mask weighted training loss in TCM improves the temporal consistency while avoiding image blurring. More visualization results are in the supplementary material.

\begin{table}[tb]
  \caption{\textbf{User studies of Real-world videos and Generation videos.}  Since the videos in these two datasets are inherently discontinuous, $E_{warp}$ cannot properly reflect temporal consistency. Here we only present the comparative results of the user study. 
  }
  \vspace{-2mm}
  \label{tab:realdata_results}
  \setlength\tabcolsep{1.75pt}
  \resizebox{\textwidth}{!}{
  \begin{tabular}{@{}lcccccccc@{}}
    \toprule
    Type & Expert& OldAnime &OldMovie & SlowMotion & TimeLapse & VideoDM & VideoLDM\\
    \midrule
    Raw($E_{warp}$)& 0.0754 & 0.1045 & 0.1099 & 0.0724 & 0.0732 & 0.1248 & 0.1601 \\
    \midrule
    Deflicker     & 38.67\% & 19.25\% & 22.83\% & 37.67\% & 31.43\% & 30.50\% & 26.34\% &\\
    Ours          & \textbf{61.33\%} & \textbf{80.75\%} & \textbf{77.17\%} & \textbf{62.33\%} & \textbf{68.57\%} & \textbf{69.50\%} & \textbf{73.67\%} &\\
  \bottomrule
  \end{tabular}
  }
  \vspace{-3mm}
\end{table}
%
\vspace{-2mm}
\subsection{Ablation Studies}\label{sec:ablation}
\begin{figure}[tb]
\begin{center}
    \includegraphics[width=1.0\linewidth]{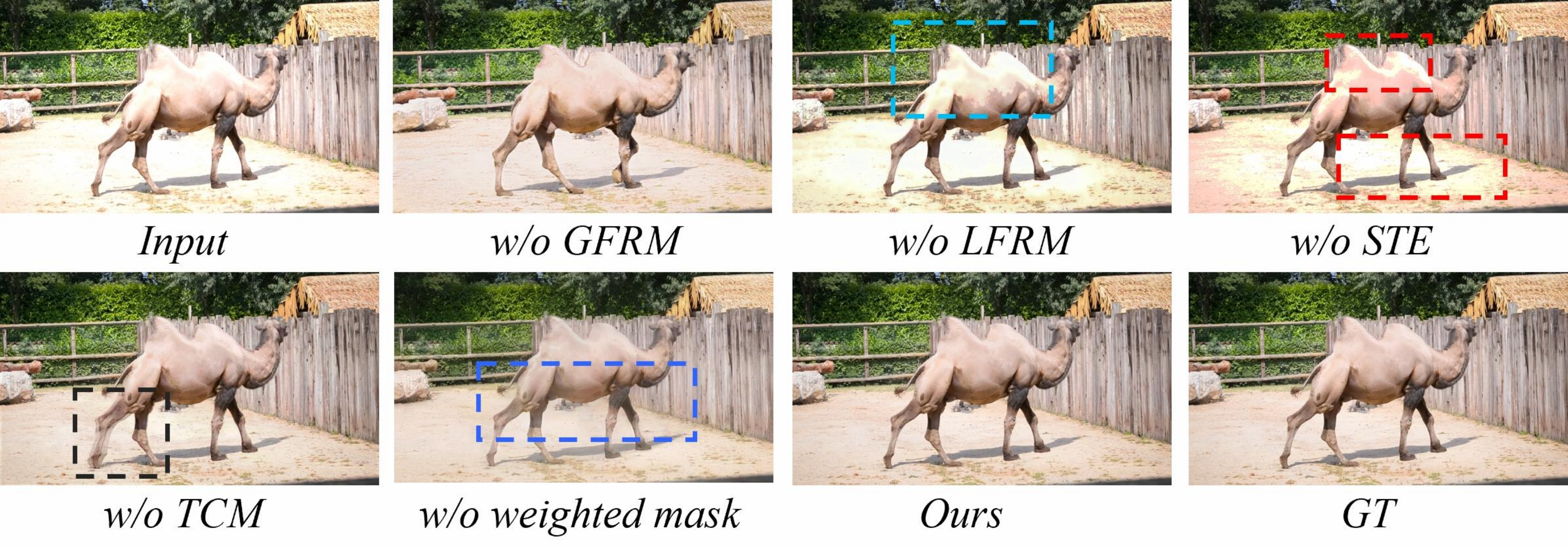}
\end{center}
    \vspace{-6mm}
   \caption{\textbf{Qualitative ablation studies of key designs in \textit{BlazeBVD}}. Specifically, local detail loss is represented in \textit{\textcolor{cyan}{cyan box}}, color artifact is represented in the \textit{\textcolor{red}{red box}}, temporal inconsistency is represented in the \textit{\textcolor{black}{black box}}, and color distortion is represented in the \textit{\textcolor{blue}{blue box}}. Zoom in for the best view.}
\label{Ablation_figure}
\vspace{-5mm}
\end{figure}

\paragraph{\textbf{STE filtering}.}
We contrast the difference between introducing STE filtering in illumination space and directly in color space. As shown in the red box of~\cref{Ablation_figure}, directly performing histogram correction in color channels separately introduces color artifacts and color distortion. Because the filtering of non-gray images in the three channels is relatively independent, the recombination after nonlinear transformation would produce such defects.
\vspace{-3mm} 
\paragraph{\textbf{GFRM and LFRM}.}
We compare the important roles of two modules, GFRM and LFRM. As shown in~\cref{tab:ablation_study}, removing GFRM and LFRM reduces PSNR and SSIM and increases $E_{warp}$ of inference videos. It can be inferred that GFRM needs to initially remove the global flicker between frames, otherwise the subsequent modules cannot maintain the temporal consistency, while LFRM plays the role of maintaining the texture fidelity of the video content, which is reflected in the cyan box of~\cref{Ablation_figure}.
\vspace{-3mm}
\paragraph{\textbf{Designed weighted loss}.}
As shown in~\cref{tab:ablation_study}, the lightweight spatio-temporal network TCM reduces $E_{warp}$, which refines the temporal consistency of the processed video. We also compare the effect of using unweighted training loss and the adaptive mask weighted loss on this temporal network.~\Cref{Ablation_figure} provides the resulting frame with unweighted training, which is generally blurred and suffers from color distortion, while the weighted training keeps the frame clear and faithful in most areas. It can be seen that adaptive mask weighting improves the temporal consistency of the inference video, improving the inter-frame continuity more pertinently.

\begin{table}[tb]
\parbox{.41\linewidth}{
\centering
\caption{\textbf{Inference time and model parameters comparison.} Our approach is faster than the baseline in inference time with less memory, and comparable in MACs and Params. 
  }
  \vspace{-2mm}
  \label{tab:time_results}
  \setlength\tabcolsep{8pt}
  \centering
  \begin{tabular}{@{}lcc@{}}
    \toprule
    Method &  Deflicker  & Ours \\
    \midrule
    Time/80p  & 614.19s & \textbf{58.37s} \\
    Memory   & 5204M & \textbf{1389M} \\
    MACs  & 260.7G & \textbf{251.4G}\\
    Params & 12.48M & 17.77M\\
    
  \bottomrule
  \end{tabular}
}
\hfill
\parbox{.55\linewidth}{
\centering
\caption{\textbf{Ablation studies in DAVIS-2017-Test.} w/o STE replaces STE with mean filtering in Stage1 and w/o weighted denotes the warping loss without mask weighting in TCM. 
  }
  \vspace{-2mm}
  \label{tab:ablation_study}
  \setlength\tabcolsep{8pt}
  \centering
  \begin{tabular}{@{}lccc@{}}
    \toprule
    Method & PSNR$\uparrow$ & SSIM$\uparrow$ & $E_{warp}\downarrow$\\
    \midrule
    w/o STE      & 26.4412 & 0.9517 & 0.1011\\
    w/o GFRM     & 23.8939 & 0.9244 & 0.1225\\
    w/o LFRM     & 27.1366 & 0.9462 & 0.1122\\
    w/o TCM      & \textbf{29.9130} & 0.9583 & 0.1093\\
    w/o weighted & 28.1049 & 0.9523 & 0.0849\\
    \midrule
    BlazeBVD     & 28.6092 & \textbf{0.9638} & \textbf{0.0825}\\
  \bottomrule
  \end{tabular}
}
\vspace{-2.5mm}
\end{table}

\vspace{-3mm}
\subsection{Discussion}
\textit{\textbf{Advantages and limitations}.}
First and foremost, BlazeBVD is over 10$\times$ faster than the inference speed of baseline using the atlas representation, which is specifically recorded in~\cref{tab:time_results}. Our method also takes advantage of explicit prior extraction in BVD task, which provides directions for improvement in subsequent work. Additionally, the cause of local flicker has been noted and effectively dealt with from both global and local perspectives. However, in the local flicker removal module, our method still has drawbacks. Due to the inaccurate optical flow motion estimation, the fusion network cannot accurately convey the local texture information of adjacent frames, and there are still slight edge artifacts and color distortion that cannot be eliminated. We hope to continue exploring the transfer and fusion of local region information between frames~\cite{Wan_2022_CVPR_videorestore}.

\noindent\textit{\textbf{Future work}.}
The proposed method can be applied to various flickering videos, which does not need to provide extra guidance, and can be directly applied to processed videos of other tasks for further refinement. During the deflickering process, we realized that the temporal consistency of the video content needed to be additionally considered. Especially in the videos obtained by the generative model, maintaining the fidelity of the content and improving the fluency are contradictory. How to better balance the degradation between faithfulness and coherence when processing videos and find reasonable metric forms is our future work.
\vspace{-2mm}
\section{Conclusion}
In this paper, we propose BlazeBVD, a universal approach for the blind video deflickering task. 
The core of our method is to prepare flicker priors within the STE filter in illumination space first, and then utilize these priors to correct the global flicker and local exposed texture. GFRM and LFRM leverage the priors to remove global and local flicker in several adjacent frames with less computation. Finally, TCM is used to improve video coherence and inter-frame consistency. 
Extensive experiments show that our method outperforms previous works on both synthetic and real-world datasets, achieving a 10$\times$ speedup in model inference, and ablation studies validate the effectiveness of designed modules. 

%
%
\bibliographystyle{splncs04}
\bibliography{main}

\begin{thebibliography}{10}
\providecommand{\url}[1]{\texttt{#1}}
\providecommand{\urlprefix}{URL }
\providecommand{\doi}[1]{https://doi.org/#1}

\bibitem{afifi2021_histogram}
Afifi, M., Brubaker, M.A., Brown, M.S.: Histogan: Controlling colors of gan-generated and real images via color histograms. In: CVPR. pp. 7941--7950 (2021)

\bibitem{flicker_free}
Anarchy, D.: Flicker free, \url{https://digitalanarchy.com/Flicker/main.html}

\bibitem{Bassiou_Kotropoulos_2007_histogram}
Bassiou, N., Kotropoulos, C.: Color image histogram equalization by absolute discounting back-off. Computer Vision and Image Understanding  \textbf{107},  108–122 (Jul 2007)

\bibitem{Bonneel_2015}
Bonneel, N., Tompkin, J., Sunkavalli, K., Sun, D., Paris, S., Pfister, H.: Blind video temporal consistency. ACM Transactions on Graphics  \textbf{34},  1–9 (Nov 2015)

\bibitem{Bottenus_Byram_Hyun_2021_matching}
Bottenus, N., Byram, B.C., Hyun, D.: Histogram matching for visual ultrasound image comparison. IEEE Transactions on Ultrasonics, Ferroelectrics, and Frequency Control  \textbf{68},  1487–1495 (May 2021)

\bibitem{Chang_2019_ICCV_adv_loss}
Chang, Y.L., Liu, Z.Y., Lee, K.Y., Hsu, W.: Free-form video inpainting with 3d gated convolution and temporal patchgan. In: ICCV. pp. 9066--9075 (2019)

\bibitem{Chu_2020_gan_video}
Chu, M., Xie, Y., Mayer, J., Leal-Taix{\'e}, L., Thuerey, N.: Learning temporal coherence via self-supervision for gan-based video generation. ACM Transactions on Graphics  \textbf{39},  75--1 (2020)

\bibitem{Delon_2006_STE}
Delon, J.: Movie and video scale-time equalization application to flicker reduction. IEEE Transactions on Image Processing  \textbf{15},  241–248 (Jan 2006)

\bibitem{Delon_Desolneux_2010}
Delon, J., Desolneux, A.: Stabilization of flicker-like effects in image sequences through local contrast correction. SIAM Journal on Imaging Sciences  \textbf{3},  703–734 (Jan 2010)

\bibitem{dosovitskiy2015flownet}
Dosovitskiy, A., Fischer, P., Ilg, E., Hausser, P., Hazirbas, C., Golkov, V., Van Der~Smagt, P., Cremers, D., Brox, T.: Flownet: Learning optical flow with convolutional networks. In: ICCV. pp. 2758--2766 (2015)

\bibitem{eslami_2024_rethinking_raft}
Eslami, N., Arefi, F., Mansourian, A.M., Kasaei, S.: Rethinking raft for efficient optical flow. arXiv preprint arXiv:2401.00833  (2024)

\bibitem{Huang_2022_ENC_exposure}
Huang, J., Liu, Y., Fu, X., Zhou, M., Wang, Y., Zhao, F., Xiong, Z.: Exposure normalization and compensation for multiple-exposure correction. In: CVPR. pp. 6043--6052 (2022)

\bibitem{Huang_2022_FECNet_exposure}
Huang, J., Liu, Y., Zhao, F., Yan, K., Zhang, J., Huang, Y., Zhou, M., Xiong, Z.: Deep fourier-based exposure correction network with spatial-frequency interaction. In: ECCV. pp. 163--180 (2022)

\bibitem{huang2023learning_exposure}
Huang, J., Zhao, F., Zhou, M., Xiao, J., Zheng, N., Zheng, K., Xiong, Z.: Learning sample relationship for exposure correction. In: CVPR. pp. 9904--9913 (2023)

\bibitem{ECLNet_2022_exposure}
Huang, J., Zhou, M., Liu, Y., Yao, M., Zhao, F., Xiong, Z.: Exposure-consistency representation learning for exposure correction. In: ACM MM. pp. 6309--6317 (2022)

\bibitem{ilg2017flownet2}
Ilg, E., Mayer, N., Saikia, T., Keuper, M., Dosovitskiy, A., Brox, T.: Flownet 2.0: Evolution of optical flow estimation with deep networks. In: CVPR. pp. 2462--2470 (2017)

\bibitem{Johnson_2016_VGG}
Johnson, J., Alahi, A., Fei-Fei, L.: Perceptual losses for real-time style transfer and super-resolution. In: ECCV. pp. 694--711 (2016)

\bibitem{Kanj_Talbot_Luparello_2017}
Kanj, A., Talbot, H., Luparello, R.R.: Flicker removal and superpixel-based motion tracking for high speed videos. In: ICIP. pp. 245--249 (2017)

\bibitem{kasten2021_atlas}
Kasten, Y., Ofri, D., Wang, O., Dekel, T.: Layered neural atlases for consistent video editing. ACM Transactions on Graphics  \textbf{40},  1--12 (2021)

\bibitem{Lai_2018}
Lai, W.S., Huang, J.B., Wang, O., Shechtman, E., Yumer, E., Yang, M.H.: Learning blind video temporal consistency. In: ECCV. pp. 170--185 (2018)

\bibitem{Lei_2023}
Lei, C., Ren, X., Zhang, Z., Chen, Q.: Blind video deflickering by neural filtering with a flawed atlas. In: CVPR. pp. 10439--10448 (2023)

\bibitem{Lei_2020}
Lei, C., Xing, Y., Chen, Q.: Blind video temporal consistency via deep video prior. In: NeurIPS. pp. 1083--1093 (2020)

\bibitem{Lin_2014_COCO}
Lin, T.Y., Maire, M., Belongie, S., Hays, J., Perona, P., Ramanan, D., Doll{\'a}r, P., Zitnick, C.L.: Microsoft coco: Common objects in context. In: ECCV. pp. 740--755 (2014)

\bibitem{Ma_2022_CVPR_exposure}
Ma, L., Ma, T., Liu, R., Fan, X., Luo, Z.: Toward fast, flexible, and robust low-light image enhancement. In: CVPR. pp. 5637--5646 (2022)

\bibitem{Mei_Patel_2023_dm}
Mei, K., Patel, V.: Vidm: Video implicit diffusion models. In: AAAI. pp. 9117--9125 (2023)

\bibitem{moniz2019luciddream_consistency}
Moniz, J.R.A., Kang, E., P{\'o}czos, B.: Luciddream: Controlled temporally-consistent deepdream on videos. arXiv preprint arXiv:1911.11960  (2019)

\bibitem{Park_2019_consistency}
Park, K., Woo, S., Kim, D., Cho, D., Kweon, I.S.: Preserving semantic and temporal consistency for unpaired video-to-video translation. In: ACM MM. pp. 1248--1257 (2019)

\bibitem{Perazzi_2016}
Perazzi, F., Pont-Tuset, J., McWilliams, B., Van~Gool, L., Gross, M., Sorkine-Hornung, A.: A benchmark dataset and evaluation methodology for video object segmentation. In: CVPR (Jun 2016)

\bibitem{perez2018perceptual_consistency}
P{\'e}rez-Pellitero, E., Sajjadi, M.S., Hirsch, M., Sch{\"o}lkopf, B.: Perceptual video super resolution with enhanced temporal consistency. arXiv preprint arXiv:1807.07930  (2018)

\bibitem{Pfeuffer_2019_temporal}
Pfeuffer, A., Dietmayer, K.: Separable convolutional lstms for faster video segmentation. In: ITSC. pp. 1072--1078 (2019)

\bibitem{Piti2003RemovingFF_tradition}
Pitie, F., Kokaram, A., Dahyot, R.: Removing flicker from old movies. Master's thesis, University of Nice-{\"E}ophia Antipolis, France, {\"E}eptember  (2002)

\bibitem{Pont-Tuset_arXiv_2017_DAVIS}
Pont-Tuset, J., Perazzi, F., Caelles, S., Arbel\'aez, P., Sorkine-Hornung, A., {Van Gool}, L.: The 2017 davis challenge on video object segmentation. arXiv:1704.00675  (2017)

\bibitem{revision}
{RE:VISION}: De:flicker, \url{https://revisionfx.com/products/deflicker/}

\bibitem{Saito_Saito_Koyama_Kobayashi_2020_gan}
Saito, M., Saito, S., Koyama, M., Kobayashi, S.: Train sparsely, generate densely: Memory-efficient unsupervised training of high-resolution temporal gan. International Journal of Computer Vision p. 2586–2606 (Nov 2020)

\bibitem{Skorokhodov_2022_gan}
Skorokhodov, I., Tulyakov, S., Elhoseiny, M.: Stylegan-v: A continuous video generator with the price, image quality and perks of stylegan2. In: CVPR. pp. 3626--3636 (2022)

\bibitem{teed_2020_raft}
Teed, Z., Deng, J.: Raft: Recurrent all-pairs field transforms for optical flow. In: ECCV. pp. 402--419 (2020)

\bibitem{Thasarathan_2019_consistency}
Thasarathan, H., Nazeri, K., Ebrahimi, M.: Automatic temporally coherent video colorization. In: CRV. pp. 189--194 (2019)

\bibitem{Thimonier_2021_consistency}
Thimonier, H., Despois, J., Kips, R., Perrot, M.: Learning long-term style preserving blind video temporal consistency. In: ICME. pp.~1--6 (2021)

\bibitem{Wan_2022_CVPR_videorestore}
Wan, Z., Zhang, B., Chen, D., Liao, J.: Bringing old films back to life. In: CVPR. pp. 17694--17703 (2022)

\bibitem{Wang_2022_LCDPNet_exposure}
Wang, H., Xu, K., Lau, R.W.: Local color distributions prior for image enhancement. In: ECCV. pp. 343--359 (2022)

\bibitem{wang2024magicvideov2_dm}
Wang, W., Liu, J., Lin, Z., Yan, J., Chen, S., Low, C., Hoang, T., Wu, J., Liew, J.H., Yan, H., et~al.: Magicvideo-v2: Multi-stage high-aesthetic video generation. arXiv preprint arXiv:2401.04468  (2024)

\bibitem{Wang_2023_DA_exposure}
Wang, Y., Peng, L., Li, L., Cao, Y., Zha, Z.J.: Decoupling-and-aggregating for image exposure correction. In: CVPR. pp. 18115--18124 (2023)

\bibitem{SSIM}
Wang, Z., Bovik, A.C., Sheikh, H.R., Simoncelli, E.P.: Image quality assessment: from error visibility to structural similarity. IEEE Transactions on Image Processing  \textbf{13},  600--612 (2004)

\bibitem{wu_2023_dm_video}
Wu, J.Z., Ge, Y., Wang, X., Lei, S.W., Gu, Y., Shi, Y., Hsu, W., Shan, Y., Qie, X., Shou, M.Z.: Tune-a-video: One-shot tuning of image diffusion models for text-to-video generation. In: ICCV. vol.~39, pp. 7623--7633 (2023)

\bibitem{Xu_2023_histogram}
Xu, R., Tu, Z., Du, Y., Dong, X., Li, J., Meng, Z., Ma, J., Bovik, A., Yu, H.: Pik-fix: Restoring and colorizing old photos. In: WACV. pp. 1724--1734 (2023)

\bibitem{Xu_2020_videowork}
Xu, X., Li, M., Sun, W., Yang, M.H.: Learning spatial and spatio-temporal pixel aggregations for image and video denoising. IEEE Transactions on Image Processing  \textbf{29},  7153--7165 (2020)

\bibitem{zhan2019_opticalflow}
Zhan, X., Pan, X., Liu, Z., Lin, D., Loy, C.C.: Self-supervised learning via conditional motion propagation. In: CVPR. pp. 1881--1889 (2019)

\bibitem{zhang2021_histogram}
Zhang, F., Shao, Y., Sun, Y., Zhu, K., Gao, C., Sang, N.: Unsupervised low-light image enhancement via histogram equalization prior. arXiv preprint arXiv:2112.01766  (2021)

\bibitem{Zhou_2020_consistency}
Zhou, Y., Xu, X., Shen, F., Gao, L., Lu, H., Shen, H.T.: Temporal denoising mask synthesis network for learning blind video temporal consistency. In: ACM MM. pp. 475--483 (2020)

\end{thebibliography}

\end{document}